\documentclass[conference]{IEEEtran}
\IEEEoverridecommandlockouts
\usepackage{cite}
\usepackage{amsmath,amssymb,amsfonts}
\usepackage{algorithmic}
\usepackage{graphicx}
\usepackage{textcomp}
\usepackage{stfloats}
\def\BibTeX{{\rm B\kern-.05em{\sc i\kern-.025em b}\kern-.08em
    T\kern-.1667em\lower.7ex\hbox{E}\kern-.125emX}}
\begin{document}
\title{Characterizing the impact of using features extracted from pre-trained models on the quality of video captioning sequence-to-sequence models \\
}

\author{\IEEEauthorblockN{Menatallh Hammad}
\IEEEauthorblockA{
\textit{Zewail City For Science And Technology}\\
Cairo, Egypt \\
s-menah.hammad@zewailcity.edu.eg}
\and
\IEEEauthorblockN{May Hammad}
\IEEEauthorblockA{
\textit{Zewail City For Science And Technology}\\
Cairo , Egypt \\
s-maymahmod@zewailcity.edu.eg}
\and
\IEEEauthorblockN{Mohamed Elshenawy}
\IEEEauthorblockA{
\textit{Zewail City For Science And Technology}\\
Cairo, Egypt \\
melshenawy@zewailcity.edu.eg}
}

\maketitle

\begin{abstract}
The task of video captioning, that is, the automatic generation of sentences describing a sequence of actions in a video, has attracted an increasing attention recently. The complex and high-dimensional representation of video data makes it difficult for a typical encoder-decoder architectures to recognize relevant features and encode them in a proper format. Video data contains different modalities that can be recognized using a mix image, scene, action and audio features. In this paper, we characterize the different features affecting video descriptions and explore the interactions among these features and how they affect the final quality of a video representation. Building on existing encoder-decoder models that utilize limited range of video information, our comparisons show how the inclusion of multi-modal video features can make a significant effect on improving the quality of generated statements. The work is of special interest to scientists and practitioners who are using sequence-to-sequence models to generate video captions.    
\end{abstract}

\begin{IEEEkeywords}
neural networks, feature extraction, machine intelligence, pattern analysis, video signal processing 
\end{IEEEkeywords}

\section{Introduction}
\noindent The problem of generating accurate descriptions for videos has received growing interest from researchers in the past few years \cite{b1} \cite{b2} \cite{b3} \cite{b4}. The different modalities contained in a video scene and the need to generate coherent and descriptive multi-sentence descriptions have imposed  several challenges on the design of effective video captioning models. Early attempts focus on representing videos as a stack of images frames such as in \cite{b5} and \cite{b6}. Such representation, however, ignores the multi-modal nature of video scenes and has a limited ability to generate accurate descriptions that describe temporal dependencies in video sequences. Also, it ignores some important video features such as activity flow, sound and emotions. This problem was addressed in more recent approaches \cite{b7} and \cite{b8}, in which videos are represented using multi-modal features that are capable of capturing more detailed information about scenes and their temporal dynamics.

While multi-modal approaches have shown promising results improving the efficiency of the video captioning techniques in general \cite{b9}\cite{b12}, their implementation has several challenges that hamper their use in video captioning applications. Challenges include:
\begin{itemize}
\item Choosing the right combination of features describing a scene: Humans have the ability to extract the most relevant information describing a movie scene. Such information describes the salient moments of a scene rather than regular dynamics (e.g. regular crowd motion in a busy street). The ability of a machine learning model to make such distinction is limited by the proper choice of features, whether 2D or 3D, that encodes the activity of each object in a scene. 
\item Choosing the right concatenation of features: presented models adopt several different concatenation techniques to merge various combinations of input feature vectors and feed them to the network as a single representation of a scene. These methods, however, do not provide detailed information about the rationale for choosing such methods and the impact of choosing the wrong concatenation method on the overall accuracy of the model.
\end{itemize}

In this paper, we study the impact of using different combinations of features from pre-trained models on the quality of video captioning model. We base our model on the state-of-the-art S2VT-encoder-decoder architecture presented in \cite{b11}. 
While the original S2VT model uses 2D RGB visual features extracted from VGG-19 network as the only descriptors for the input video frames, we compared the performance of the model using different combinations of rich video features that describe various modalities  such as motion  dynamics,  events and sound. Additionally, We have added an attention-based model to improve the overall quality of the decoder. All video features are extracted using pre-trained models and saved in a combined representation as to decrease the complexity of computations at training time. The encoder network was fed the combine video frame feature vectors and used to construct corresponding representation. This representation is fed to the decoder network to construct the text.

\section{RELATED WORK}
Video captioning methods rely on two basic approaches to generate video descriptions: template-based and sequencing models \cite{b13}. The first approach, template-based models, generate video descriptions using a set of predefined rules and templates that construct natural language sentences \cite {b14}\cite{b15}. The visual features of the video  are identified and utilized to generate semantic representations that can be encoded and utilized withing the predefined templates. While template-based models generally produce more robust video descriptions, they have limited ability to represent semantically-rich domains which incorporate large number of entities, sophisticated structures and complex relationships. The effort to construct rules and templates for such domains is  prohibitively expensive \cite{b14}.\\ 

 The second approach is to construct sequence to sequence models that map video features representation into a chain of words. Inspired by  recent advancements in computer vision and neural machine translation techniques, these models employ an encoder-decoder framework in which various sets of video features, typically extracted from pre-trained 2D or 3D convolutional neural networks (CNN), are fed to a neural network model. These sets are encoded to create a representation of the video content, and then decoded to produce a sequence of words summarizing this representation. For example,  Venugopalan et al.\cite{b11}, presented  a long short-term memory (LSTM) model that encodes visual features from video frames to generate video descriptions. At the encoding stage, the model combines features generated from a CNN at frame level and uses them to generate a feature vector representing the video content. The resulting vector is then fed to an LSTM decoder, at the decoding stage, to generate the corresponding video caption. \\

Recent research efforts have focused on examining different video features to improve the accuracy of sequencing video captioning models.  Yao et al. \cite{b16} focus on using a spatial temporal 3-D convolutional neural network (3-D CNN) to capture short temporal dynamics within videos. Their approach utilizes an attention mechanism that gives different weights to video frames during the decoding stage. In addition visual features, audio features have also been investigated to improve the accuracy of the sequencing models. Ramanishka et al. \cite {b17} proposed a multi-modal model approach  which incorporates visual features, audio features and video topic to produce a video representations. Their model uses  Mel Frequency Cepstral Coefficients (MFCCs) of the audio stream to represent the audio features.  Jin et al. \cite{b18} also propose a multi-modal approach which combines different video modalities including visual, audio, and meta-data. The multi-modal approach was also employed by Long et al. \cite{b19} and Hori et al.\cite{b20}, who utilize  attention mechanisms to extract the most salient visual and audio features and use them to generate video descriptions. \\ 

Although the above-mentioned model have demonstrated effective methods to combine multi-modal video features, there is a lack of studies that explores which visual, audio and scene features can represent the video content better. In this paper, we focus on features extracted from pre-trained networks providing a comparative assessment of the impact of choosing these features on the overall quality of the model.

\section{ Approach} 
As mentioned earlier, the model used in this paper was based on the S2VT-encoder-decoder architecture \cite{b11}.  Training the model involves the following three stages:
\begin{itemize}
\item
The first stage involves the conversion of input video frames into high level descriptive feature vectors describing the video content. Generated feature vectors are used as input to the second phase (the encoding phase) .
\item 
In the encoding stage, an LSTM based encoder is utilized to  encode the feature vectors generated from the first stage  into a latent representation describing the given video scene.

\item
The third stage is the decoding stage, in which the representation generated from the LSTM encoder is fed to the LSTM decoder to produce a chain of words representing the video description.
\end{itemize}
In the following subsection, we describe these stages in more details providing information of how the basic architecture was extended to test different modalities of the video scene. We start, in subsection A, by discussing the details of the S2VT Model used in our experiments highlighting its underlying operation principles. Then, we discuss the type of information and selected input frame features that were fed to the S2VT model.

\subsection{Basic Operation of the S2VT Model }

The S2VT  or Sequence to Sequence - Video to Text model \cite{b11} is a sequence to sequence encoder-decoder model. The model takes, as input, a sequence of video frames
$(x 1 , . . . , x n )$, and produces a sequence of words
$(y 1 , . . . , y m )$. Both input and output are allowed to have variable sizes. The model tries to maximize the conditional probability of producing output sentence give the input video frames, that is: 

$$p(y 1 , . . . , y m |x 1 , . . . , x n )$$

To do so, the S2VT model relies on using an LSTM model
for both the encoding and decoding stage. In our experiments, we used a stack of two LSTMs with 512 hidden units each, where the first layer produces the video representation, and the second layer produces the output word sequence.

Similar to original design, which is explained in more detail in \cite{b11}, the top LSTM layer receives a sequence of input video frames and encodes them to produce a latent representation of the video content ($h_{t}$). No loss is calculated in the encoding stage. 
When all the frames are encoded, the second LSTM is fed the beginning-of-sentence tag $BOS$, which triggers it to start decoding and producing the corresponding sequence of words.

During the decoding stage, the model maximizes the log-likelihood of the predicted output word given the hidden representation produced by the encoder, and given the previous words produced. That is, produced words rely on the hidden representation and previous generated words.

For a model with parameters $\theta$ The maximization equation can be expressed as:
$$\theta ^{*}=argmax(\sum_{t=1}^{m}\log (p(y t |h_{ n+t−1} , y_{ t-1} ; \theta) ))$$
Optimization occurs over the entire dataset using gradient descent. During optimization, the loss is back-propagated to allow for a better latent representation ($h_{t}$). An end-of-statement tag $EOS$ is used to indicate that that the sentence is terminated.

\subsection{Selected Multi-modal Input Video Frame Features  }
In our experiments, we investigate the impact of the following features extracted from pre-trained models. We examined five type of information and their impact on the quality of the S2VT model. The following list summarizes the category of features and the models used to generate them:
   \begin{itemize}
        \item 2D object recognition Features: As the name indicates, these features relies on the 2D characteristics of the video frames. They act to provide information about different objects in the scene and how the existence/ non-existence of objects impact the video description. To get these features, we used pre-trained ResNext-101(32*16d) model \cite{b21}. The ResNext models are pre-trained in weakly-supervised fashion on 940 million public images with 1.5K hashtags matching with 1000 ImageNet1K synsets.
        
        \item 2D object recognition Intermediate Features: Intermediate features provides more abstract information about the different objects in a frame and how they are related. These features are applied in near duplicate video retrieval \cite{b26}, and there are generally used to provide more comprehensive information about the objects in each frame.

        \item Scene Recognition Features: These features are used to present a descriptive view of each video scene.  Scene information was provided using a ResNet50 network \cite{b22} trained on Places365 dataset\cite{b23}. 

        \item   3D Action Recognition Features: These features are used to provide information about the activities of the different objects detected in the movie scene. We used the kinetics-i3d model \cite{b24}, which is a CNN model that is pre-trained on Kinetics dataset (contains around 700 action classes).
        
        \item Audio features: Audio features are used to provide information about the different sounds in the video to use it in generating descriptions. In our experiments, we used SoundNet proposed in \cite{b25}, which is used to recognize objects and scenes from sounds. The network is pre-trained using the visual features from over 2,000,000 videos. It uses a student-teacher training procedure that transfers visual knowledge from visual models such as ImageNet into sound modality.Therefore, it helps associate sounds with scene and object recognition tasks. We extracted sound features from layer 24 of the model.

\end{itemize}

\begin{figure}[tbp]
\centerline{\includegraphics[width=\linewidth]{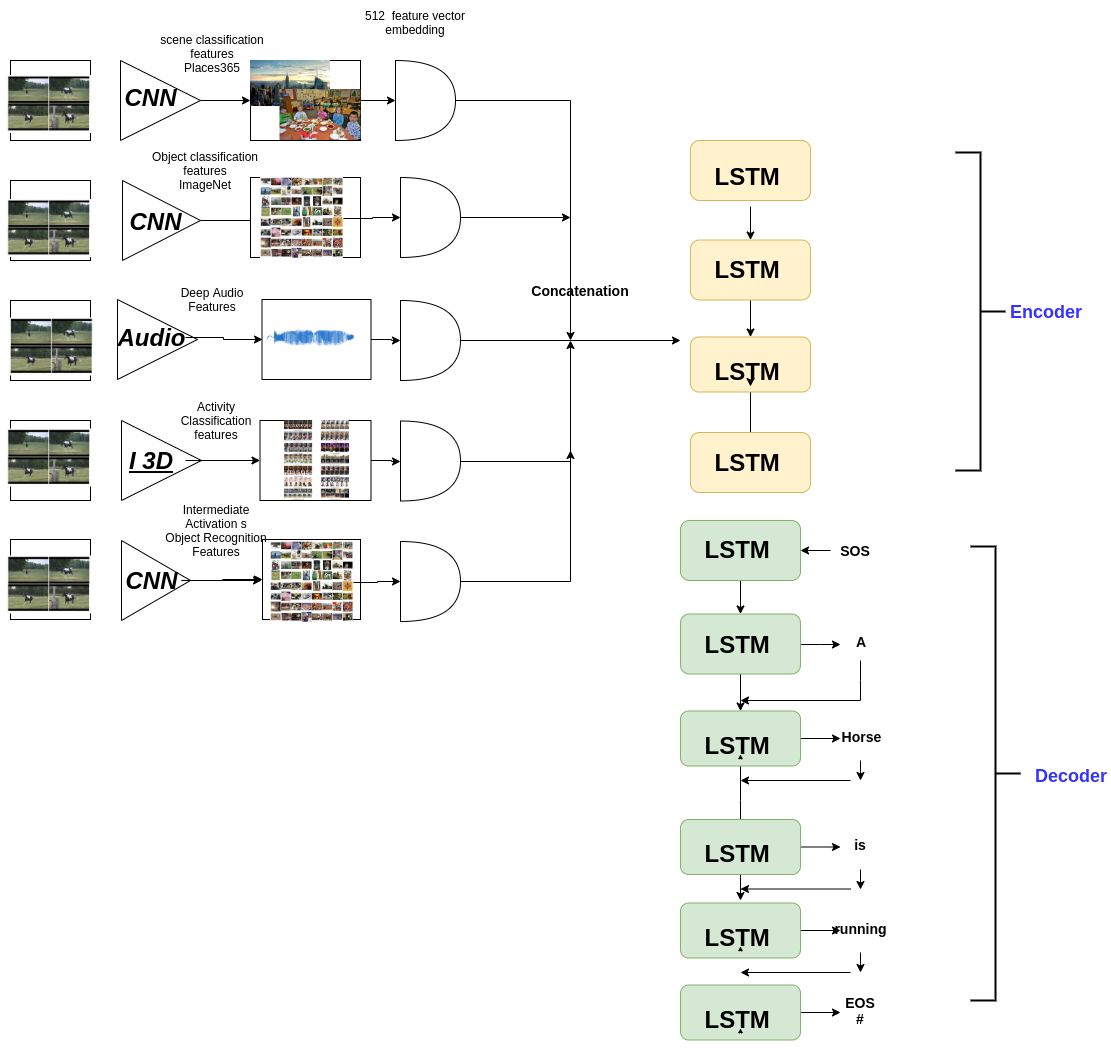}}
\caption{The figure describes our proposed approach for the video captioning task. It extends the S2VT state of the art model and exploits a multi modal video representation to best understand the video content. The input video features are encoded using LSTM encoder, then accordingly the encoder hidden states are inputs to the decoder that generates a caption that summarizes the representation. We used a combination of the 2D object features, scene feature, activity features and sound features extracted from the video as an input to the network. }
\label{fig}
\end{figure}

.

\section{Experimental Setup }

\subsection{DATASETS}\label{AA}
The model was trained and evaluated using the Microsoft Research Video to Text (MSR-VTT) \cite{b27}. MSR-VTT is a   video  description  dataset that includes 10,000 video clips  divided into 20 categories of video content such as  music, sports, gaming, and news. Each clip is annotated with 20 natural language sentences, which provides a rich set for training and validation. When training the model, we followed the original split in MSR-VTT, we split the data according to 65\%:30\%:5\%, for the training, testing and validation sets respectively,  which corresponds to 6,513, 2,990 and 497 clips for each set. 8,810 videos out of the 10000 clips has audio content. The audio content of these silent clips is replaced by zeros.

\subsection{Implementation}
\subsubsection{Feature extraction}
 As discussed earlier, we extract five types of video features: the 2D object recognition visual features using the ResNext101(32x16D) pre-trained network; 2D object recognition Intermediate Features using a pre-trained VGG-19 network; scene recognition features using a ResNet50 network trained on Places365 dataset; action recognition features using I3D pertained network; and deep sound features using the pre-trained SoundNet. These features are used, as we will discuss in the next section, to conduct several comparisons that characterize the impact of each video information type on the accuracy of the model. The following paragraphs provide more details about the feature extraction process:

\begin{itemize}
\item The 2D object recognition visual features: Input videos are sampled into 40 evenly-spaced frames, if the video instance contains less than 40 frames, we pad zero frames at the end of original frames. The frames are fed to the pre-trained ResNext101 (32x16d) network.Features are extracted directly from the last fully connected layer (FC7) following the maxpool layer. We Followed the pre-processing of inputs for ResNxt 32x16d. The height and width of each frame of the video were resized to  256, to then use the central crop patch of 224x224 pixels of each frame. A normalisation of the pixel using mean and standard deviation (mean=[0.485, 0.456, 0.406], std=[0.229, 0.224, 0.225]) are  applied.
\item 2D object recognition Intermediate Features: Similar to \cite{b26}, a sampling rate of 1 frame per second was used, and features are extracted using VGG-16 from the intermediate convolutional layers. To generate a single frame descriptor, the Maximum Activation of Convolutions (MAC) function is applied on the activation of each layer before concatenating them to produce a single frame descriptor.
\item Scene Recognition Features:  we extracted the scene description from the 2048 dimensional feature vector from last fully connected layer after the mean-pool layer of the Resnet50.
\item 3D Action Recognition Features: As discussed earlier, we utilized the pre-trained I3D model initially trained  on Kinetics dataset for human action recognition. We extracted the 3D motion and activity features from the 1024 dimensional vector form  fully-connected layer (FC6). The resulting extracted feature size is 1024 per video frame.
\item Audio features: For extracting the associated audio features we used the state-of-the-art SoundNet architecture which represents a CNN pretrained on AudioSet. We split the videos into both audio wav files and video content. The audio files are used as input to the SoundNet model, then the output feature vector is extracted from  24th convolutional layer of the network which provides 1024 dimensional feature vector for the input audio file.
\end{itemize}

\subsubsection{Description Sentences Processing}
We started by converting all sentences to lower case and removing all the punctuation. Then the sentences were split up using blank spaces to generate the  chain of words used in the modelling procedure. Both $BOS$ and $EOS$ tags were added to the begging and end of the annotation sentence with a max length of 28 words. To input words, we employ a 512 dimensional embedding layer that convert words to vector representations.

\subsubsection{Optimization and regularization parameters}

We utilized an Adam optimizer with a learning rate of $10^{-4}$ and batch size of 128. An LSTM's dropout percentage of 0.2 was employed as a regularization technique.
\subsection{Evaluation}
 We used four main metrics to evaluate the quality of the sequence model: 
 \begin{itemize}
 \item BLEU score \cite {b28}, which is a standard technique to compare an output statement to its corresponding reference descriptions in the dataset. In our analysis, we used four n-gram BLEU scores of sizes 1 to 4.  
 \item METEOR score \cite{b29}, which, in contrast to the BLEU score that uses precision-based metrics, emphasizes recall in addition to precision when comparing generated statements to their reference descriptions.
\item  CIDEr \cite{b30}, which is  tailored for image description tasks and uses human consensus to evaluate generated description. In our analysis, we used the standard CIDEr-D version, which is available as part of the Microsoft COCO evaluation server \cite{b31}. 
\item ROUGE-L \cite{b32}, which is tailored to evaluate summarization tasks. The metric rely on assessing the number of overlapping units between the statements generated by the model and reference statements. These units include n-gram, word sequences, and word pairs. 
 \end{itemize}

\section{RESULTS AND ANALYSIS}

We conducted several experiments using the five video features discussed earlier. The results are shown in table \ref{table:results}. In the first set of experiments, we tested the impact of using each feature type when it is used separately to train the model. We found that the 2D object  recognition visual features acquired using the pre-traing ResNext101 network achieved the best results compared to other feature types in BLEU1, BLEU2, BLEU3, BLUE4, METEOR, and CIDEr scores. The 3D action recognition features, acquired form the pre-trained I3D model, achieved the best score in ROUGE-L score.

In the second set of experiments, we tried different combinations of pre-trained features. First, we tried detailed visual features only (2D object recognition visual features, scene  recognition features, and 3D action recognition features). Audio features and 2D object recognition intermediate features were added in the subsequent experiments as shown in table \ref{table:results}.The combination of 2D object recognition visual features, scene  recognition features, and 3D action recognition features achieved the best results in BLEU1, BLEU2, BLEU3, BLUE4, METEOR, and ROUGE-L scores. Combining audio features with these features results in a slight improvement in CIDEr score. The results of using a combination of features exceed the performance of using single features in the majority of the scores.

\begin{table*}[tbp]
\caption{Results}
\centering
 \begin{tabular}{||c c c c c c c c||} 
 \hline
 FEATURES & BLEU1 & BLEU2 & BLEU3 & BLEU4 & METEOR & CIDEr & ROUGE-L \\ [0.5ex] 
 \hline
 ResNext50 (place365) & 0.77132 & 0.61591 & 0.47293 & 0.34740 & 0.26163 & 0.38299 &0.57909 \\ 
 \hline
 ResNext101 &  0.79972 & 0.65690 & 0.51847 & 0.39288 & 0.27851 & 0.59977 & 0.47940\\
 \hline
 
 SoundNet features & 0.68704 & 0.50683 & 0.37306 &0.25742 & 0.20562 &  0.51485 & 0.13247\\
 \hline
 Intermediate features & 0.76530 & 0.61260 & 0.47171 & 0.35078 & 0.26223 & 0.37009 &  0.57675\\
 \hline
 I3D(Kinetics) & 0.79185 & 0.63785 & 0.49557 & 0.37054 & 0.27282 & 0.41092 & 0.59103\\

 \hline
 place365, I3D, ResNext101  & 0.81020 & 0.66874& 0.53141 & 0.40756 & 0.28018 & 0.46732 & 0.60699\\
  \hline

 ResNext101, SoundNet, place365, I3D  &  0.80441 & 0.65539 & 0.51480 & 0.39106 & 0.28262 & 0.59949 & 0.47938\\
 \hline
 Intermediate, ResNext101, SoundNet, place365, I3D  &0.80784 & 0.65840 & 0.51841 & 0.39274 & 0.27811 & 0.45747 & 0.59797  \\
 \hline
 
\end{tabular}
\label{table:results}
\end{table*}

\section{Conclusion}

The implementation of efficient multi-modal video involve the use of several features to represent objects, actions, scenes and sounds. These features help the model extract the most pertinent information related to the scene and use this information to generate appropriate text. In this paper, we focus on the characterization of the impact of using features from pre-trained model to implement video captioning sequence-to-sequence models. Our approach  is based on the Sequence to Sequence - Video to Text (S2VT) model which uses an encoder-decoder structure to generate video descriptions.

In our experiments, we used five types of features: 2D object recognition features, extracted from a ResNext network; scene recognition features, extracted a ResNet50 network that is trained on places 365 dataset; 3D action recognition features, extracted from kinetics-i3d model, audio features, extracted using soundnet; and object recognition Intermediate Features, which provides more abstract information about the different objects in the frame and their relations. We conducted two sets of experiments, in which we examined the impact of each feature type separated form the others and the impact of using different combinations of features.

We used four main metrics for comparison: BLEU, METEOR, SIDEr, and ROUFE-L. Our results, when using a single feature type as an input to the mode, indicate that the use of 2D object recognition features achieved best score among the other feature types. When using a combination of features, we found that 2D object recognition visual features, scene  recognition features, and 3D action recognition features are the most relevant. The performance of the model when using a multi-modal approach exceeds the performance when using a single feature type for the majority of cases.

Our work can be extended in many ways. First, we would like to investigate different concatenation techniques that reduces the size of the multi-modal input data, and examine the impact of using these techniques on improving the overall efficiency of the model. Second, we would like to increase the capacity of our model, by adding more hidden nodes to the LSTM network and better regularization techniques, and examine the impact of such increase on the efficiency and accuracy of the model. Finally, we would like to investigate better modelling techniques that allow describing, summarizing and linking the salient moments in related videos.

\end{document}